\documentclass[letterpaper]{article} % DO NOT CHANGE THIS
\usepackage{aaai2026}
\usepackage{times}  % DO NOT CHANGE THIS
\usepackage{helvet}  % DO NOT CHANGE THIS
\usepackage{courier}  % DO NOT CHANGE THIS
\usepackage[hyphens]{url}  % DO NOT CHANGE THIS
\usepackage{graphicx} % DO NOT CHANGE THIS
\usepackage{natbib}  % DO NOT CHANGE THIS AND DO NOT ADD ANY OPTIONS TO IT
\usepackage{caption} % DO NOT CHANGE THIS AND DO NOT ADD ANY OPTIONS TO IT
\usepackage{algorithm}
\usepackage{algorithmic}
\usepackage{amsmath}
\usepackage{amssymb}
\usepackage{booktabs}
\usepackage{multirow}
\usepackage{newfloat}
\usepackage{listings}
\DeclareCaptionStyle{ruled}{labelfont=normalfont,labelsep=colon,strut=off} % DO NOT CHANGE THIS
\floatstyle{ruled}
\newfloat{listing}{tb}{lst}{}
\floatname{listing}{Listing}

\title{The Sound of Risk: A Multimodal Physics-Informed Acoustic Model for Forecasting Market Volatility and Enhancing Market Interpretability}

\author{
	Xiaoliang Chen*, Xin Yu, Le Chang, Teng Jing, Jiashuai He, Ze Wang, Yangjun Luo, Xingyu Chen, Jiayue Liang, Yuchen Wang, Jiaying Xie
}
\affiliations{
	SoundAI Technology, Beijing, China\\
	*Corresponding Author: chenxiaoliang@soundai.com
}

\begin{document}
	\maketitle
	
	\begin{abstract}
		Information asymmetry in financial markets, often amplified by strategically crafted corporate narratives, undermines the effectiveness of conventional textual analysis. We propose a novel \emph{multimodal} framework for financial risk assessment that integrates textual sentiment with paralinguistic cues derived from executive vocal tract dynamics in earnings calls. Central to this framework is the \textbf{Physics-Informed Acoustic Model (PIAM)}, which applies nonlinear acoustics to robustly extract emotional signatures from raw teleconference sound subject to distortions such as signal clipping. Both acoustic and textual emotional states are projected onto an interpretable three-dimensional \emph{Affective State Label} (ASL) space—\emph{Tension}, \emph{Stability}, and \emph{Arousal}. Using a dataset of 1,795 earnings calls (\(\approx\)1,800 hours), we construct features capturing dynamic shifts in executive affect between scripted presentation and spontaneous Q\&A exchanges. Our key finding reveals a pronounced divergence in predictive capacity: while multimodal features do not forecast directional stock returns, they explain up to \textbf{43.8\%} of the out-of-sample variance in 30-day realized volatility. Importantly, volatility predictions are strongly driven by emotional dynamics during executive transitions from scripted to spontaneous speech, particularly reduced textual stability and heightened acoustic instability from CFOs, and significant arousal variability from CEOs. An ablation study confirms that our multimodal approach substantially outperforms a financials-only baseline, underscoring the complementary contributions of acoustic and textual modalities. By decoding latent markers of uncertainty from verifiable biometric signals, our methodology provides investors and regulators a powerful tool for enhancing market interpretability and identifying hidden corporate uncertainty.
	\end{abstract}

	\section{Introduction}
	The efficiency and fairness of financial markets are contingent upon the transparent flow of information. Yet, a persistent asymmetry of information between corporate insiders and external stakeholders can undermine investor confidence and distort market outcomes. Quarterly earnings calls, a cornerstone of corporate disclosure, are intended to bridge this information gap. However, they often feature meticulously scripted narratives designed to manage perceptions and potentially obscure underlying risks. Consequently, prevailing analytical methods that rely solely on textual content, even those powered by advanced Large Language Models (LLMs), remain vulnerable to such narrative engineering, leaving investors at a disadvantage \citep{li2023large}.
	
	This study proposes a paradigm shift in financial risk assessment by moving beyond what is said to how it is said. We scrutinize two complementary channels: the paralinguistic properties of vocal tract dynamics and the emotional sentiment of the transcribed text. Our central postulate, grounded in dynamical systems theory and computational paralinguistics, is that the cognitive load associated with managing uncertainty or concealing information correlates with involuntary, chaotic perturbations in vocal tract dynamics \citep{schuller2013computational}. These vocal tract dynamics are physiologically rooted and thus more difficult to consciously control than word choice.
	
	To operationalize this hypothesis, we introduce the Physics-Informed Acoustic Model (PIAM), a framework engineered for robust, real-time analysis of executive sound. The model's unique strength lies in its ability to holistically process a single sound stream to simultaneously generate a transcript, classify vocal emotion, and detect acoustic events. This integrated, multi-output nature distinguishes it from conventional pipelines that typically handle these tasks in isolation. While the broader task of forecasting market reactions from earnings calls has been explored, our holistic, physics-informed approach represents a significant departure from prior methods that often rely on separate models for speech and text or use handcrafted acoustic features. Furthermore, its foundation in nonlinear acoustics makes it exceptionally well-suited to the complex acoustic environments of corporate communications. The very nonlinear distortions introduced by teleconferencing systems—such as signal clipping and compression artifacts—are phenomena that our physics-informed approach is inherently designed to model, allowing for more robust analysis than methods that treat such distortions as generic noise.
	
	To create a unified analytical framework, we map the discrete emotion labels—from both PIAM's acoustic analysis and a state-of-the-art LLM applied to the transcripts—onto a predefined, interpretable three-dimensional space. We term this the Affective State Label (ASL) space, characterized by Tension, Stability, and Arousal. Leveraging a large-scale corpus of earnings calls, we extract ASL-derived features that capture not only the statistical moments of these states but also their dynamic shifts between the scripted presentation and the unscripted Q\&A session. Our empirical investigation yields a key insight: emotional cues, whether from vocal or textual channels, primarily signal impending uncertainty rather than directional price movements. While these features are ineffective for forecasting Cumulative Abnormal Returns (CAR), they demonstrate potent predictive capability for future realized volatility. This finding suggests that an executive's expressed emotional state encodes latent corporate risks that may be absent from purely semantic disclosures or historical price data. By providing a more direct and less manipulable signal of risk, this methodology offers a valuable tool for augmenting market transparency and fostering more resilient financial ecosystems.
	
	\section{Related Work}
	Our research is situated at the intersection of financial technology, affective computing, neural networks, and acoustic signal processing.
	
	\subsection{Alternative Data in Finance}
	The quest for an informational advantage has spurred the integration of "alternative data" into financial modeling. Textual data from news, social media, and regulatory filings have been a primary focus, with LLMs enabling sophisticated sentiment analysis and event extraction \citep{li2023large}. However, the inherent manipulability of strategically released data highlights a need for more robust, less easily controlled information channels. Our work contributes to this search by exploring the acoustic modality alongside advanced textual analysis. We propose a dual-channel approach that leverages both a physiologically grounded signal (acoustics) and a deep semantic signal (text). This multimodal philosophy posits that combining information from different sources enhances robustness and provides a more holistic understanding. It finds parallels in fields like human-robot interaction, where multimodal inputs are important for interpreting user states accurately \citep{su2023recent}.
	
	\subsection{Computational Paralinguistics and Affective Computing}
	The field of computational paralinguistics has established that paralinguistic properties are rich carriers of information about a speaker's emotional, cognitive, and even physiological state \citep{schuller2013computational}. Early explorations in finance attempted to leverage these cues, demonstrating that vocal parameters could hold predictive power for market outcomes \citep{mayew2013voice}. However, these pioneering studies often relied on large sets of hand-crafted, Low-level Acoustic Descriptors (LLDs) such as pitch contours, intensity, and Mel-Frequency Cepstral Coefficients (MFCCs). While foundational, these methods face key limitations in real-world settings. Earnings call sound is often plagued by low bitrates, background noise, and reverberation, challenges extensively documented in speech enhancement literature \citep{loizou2007speech}. These factors, along with complexities from single-channel recordings \citep{benesty2008microphone}, often limited the efficacy of traditional LLD-based models. In contrast, we advance this line of inquiry by adopting an end-to-end deep learning approach. Our PIAM model learns relevant representations directly from raw sound, bypassing manual feature engineering and enabling it to capture subtle, high-dimensional patterns that handcrafted features might miss.
	
	\subsection{Advanced Acoustic Algorithms and Physics-Informed Models}
	Recent breakthroughs in sound processing have been driven by self-supervised learning. Models like wav2vec 2.0 \citep{baevski2020wav2vec} and architectures such as the Conformer \citep{gulati2020conformer} learn powerful representations from vast amounts of unlabeled sound data, forming the backbone of modern speech systems. This has culminated in the development of large-scale foundation models for both general sound and voice understanding \citep{an2024funsoundllm}. Concurrently, Physics-Informed Neural Networks (PINNs) have demonstrated the value of embedding domain-specific physical laws as inductive biases, leading to improved generalization, data efficiency, and interpretability, particularly in scientific domains \citep{karniadakis2021physics}. Our PIAM model uniquely synergizes these two frontiers. It leverages a powerful self-supervised encoder while incorporating principles from nonlinear acoustics to ensure the learned vocal state manifold is physically plausible. This fusion of data-driven learning with first-principles knowledge echoes interdisciplinary advances in scientific machine learning \citep{chen2025synergistic} and represents a principled approach to designing models for complex, noisy, and physically-grounded phenomena.
	
	\section{The PIAM Model and Feature Engineering Pipeline}
	Our methodology is a multi-stage pipeline designed to transform raw sound from earnings calls into interpretable vocal and textual dynamic features suitable for financial prediction tasks.
	
	\subsection{Physics of Vocal Dynamics}
	Human vocal production is an inherently nonlinear dynamical system. Under cognitive or emotional stress, the vocal folds can transition from stable, periodic vibrations to more complex, aperiodic, or even chaotic regimes. These transitions manifest as audible changes in vocal tract dynamics, such as roughness or subharmonics. As foundational texts on acoustics describe, linear models are ill-equipped to capture these essential phenomena \citep{blackstock2000fundamentals}. Our work is informed by the principles of nonlinear acoustics, which specifically model how high-amplitude sound waves behave \citep{hamilton2024nonlinear}. The Westervelt equation, a fundamental model in this field, serves as the physical prior for our model:
	\begin{equation}
		\nabla^2 p - \frac{1}{c_0^2}\frac{\partial^2 p}{\partial t^2} = - \frac{\beta}{\rho_0 c_0^4} \frac{\partial^2 p^2}{\partial t^2},
		\label{eq:westervelt}
	\end{equation}
	where $p$ is the sound pressure, $c_0$ is the small-signal sound speed, $\rho_0$ is the ambient density, and $\beta$ is the coefficient of nonlinearity. This equation captures how pressure itself affects the speed of sound, leading to waveform distortion and the generation of higher harmonics—the very signatures of vocal fold strain. We use the lossless form of the Westervelt equation here, focusing on nonlinear effects without dissipation.
	
	\subsection{Physics-Informed Regularization}
	A core challenge in analyzing earnings call sound is the presence of signal distortions endemic to teleconferencing systems. These are not merely random noise but systematic, \textit{nonlinear transformations} of the original vocal signal. Key artifacts include:
	\begin{itemize}
		\item \textbf{Microphone Clipping:} Occurs when an executive speaks too loudly or too close to the microphone, causing the waveform to be abruptly flattened. This is a classic nonlinear distortion that introduces spurious higher-order harmonics.
		\item \textbf{Compression Artifacts:} Low-bitrate codecs (e.g., G.711, Speex) used in telephony introduce quantization errors and other nonlinearities to reduce bandwidth.
	\end{itemize}
	These phenomena are precisely what the principles of nonlinear acoustics, as described by the Westervelt equation (Eq. \ref{eq:westervelt}), are designed to model. Therefore, instead of treating these distortions as generic noise to be filtered out, we propose a principled approach: using the governing physical equation as an inductive bias to regularize our model's latent space.
	
	The physics-informed regularizer, $\mathcal{L}_{\text{phys}}$, is defined as the mean squared residual of the Partial Differential Equation (PDE), approximated by a neural network operator $\mathcal{F}_{\theta}$:
	\begin{equation}
		\mathcal{L}_{\text{phys}} = \frac{1}{T} \sum_{t=1}^{T} \left\| \mathcal{F}_{\theta}(\mathbf{h}_t) \right\|_2^2
		\label{eq:phys_loss_revised}
	\end{equation}
	Here, $\mathbf{h}_t$ is the latent representation at time step $t$. The operator $\mathcal{F}_{\theta}$ is a shallow Multi-Layer Perceptron (MLP) that maps $\mathbf{h}_t$ to a scalar representing the sound pressure $p(t)$. Using automatic differentiation \citep{raissi2019physics}, we compute the necessary temporal derivatives of $p(t)$ to evaluate the PDE residual. This loss term penalizes latent space trajectories that are inconsistent with the physics of nonlinear sound propagation. By doing so, it encourages PIAM to learn representations that are inherently robust to the specific nonlinear distortions found in its target environment—telephone conferences.
	
	The total loss for fine-tuning becomes a weighted sum of the task losses and this physics-based regularizer:
	\begin{equation}
		\mathcal{L}_{\text{total}} = \mathcal{L}_{\text{task}} + \lambda \mathcal{L}_{\text{phys}}
	\end{equation}
	where $\lambda$ is a hyperparameter balancing task performance and physical consistency.
	
	\subsection{Model Architecture and Training}
	As outlined in Algorithm \ref{alg:piam}, the PIAM architecture transforms a raw sound signal into transcribed text and a vocal emotion label.
	\begin{enumerate}
		\item \textbf{Self-Supervised Encoder}: A pre-trained wav2vec 2.0-style encoder forms the foundation, learning powerful, contextualized representations directly from raw sound, using a contrastive loss $\mathcal{L}_{\text{NCE}}$ (Noise Contrastive Estimation).
		\item \textbf{Attention-Based Temporal Modeler}: A Bi-directional LSTM (Bi-LSTM) with an attention mechanism is added on top of the encoder's outputs. This captures long-term dynamics and, inspired by the transformer architecture's success \citep{vaswani2017attention}, focuses on the most salient sound segments for an utterance-level representation.
		\item \textbf{Multi-Task Output Head}: The final context vector is fed into a multi-task head for simultaneous Connectionist Temporal Classification (CTC)-based transcription and discrete emotion classification.
	\end{enumerate}
	
	\begin{algorithm}[tb!]
		\caption{PIAM: Acoustic Emotion and Textual Features Modeling from Raw Sound}
		\label{alg:piam}
		\textbf{Input}: Raw sound waveform segment $X$ \\
		\textbf{Output}: Emotion $e$, Text $T_{text}$
		\begin{algorithmic}[1]
			\STATE {\# 1. Self-Supervised Encoding (Pre-training)}
			\STATE $\mathbf{Z} \leftarrow \text{EncoderCNN}(X)$
			\STATE $\mathbf{C} \leftarrow \text{Quantizer}(\mathbf{Z})$
			\STATE $\mathbf{H} \leftarrow \text{TransformerContext}(\text{Mask}(\mathbf{Z}))$
			\STATE Pre-train model using contrastive loss $\mathcal{L}_{\text{NCE}}$.
			\STATE {\# 2. Temporal Aggregation (Fine-tuning)}
			\STATE $\mathbf{h}_{1..T} \leftarrow \text{FineTunedEncoder}(X)$
			\STATE $\mathbf{c}_{1..T} \leftarrow \text{Bi-LSTM}(\mathbf{h}_{1..T})$
			\STATE $\alpha_{1..T} \leftarrow \text{Attention}(\mathbf{c}_{1..T})$
			\STATE $\mathbf{v}_{\text{utterance}} \leftarrow \sum_{t=1}^{T} \alpha_t \mathbf{c}_t$
			\STATE {\# 3. Physics-Informed Multi-Task Output Head}
			\STATE $P_{\text{emotion}} \leftarrow \text{softmax}(\text{HeadEmotion}(\mathbf{v}_{\text{utterance}}))$
			\STATE $e \leftarrow \text{argmax}(P_{\text{emotion}})$
			\STATE $T_{text} \leftarrow \text{CTCDecoder}(\text{HeadTranscription}(\mathbf{h}_{1..T}))$
			\STATE Compute $\mathcal{L}_{\text{task}}$ and $\mathcal{L}_{\text{phys}}$ (from Eq. \ref{eq:phys_loss_revised}).
			\STATE Fine-tune with $\mathcal{L}_{\text{total}} = \mathcal{L}_{\text{task}} + \lambda \mathcal{L}_{\text{phys}}$.
			\STATE \textbf{return} $e, T_{text}$
		\end{algorithmic}
	\end{algorithm}
	
	\subsection{Feature Engineering via ASL Space Mapping}
	The PIAM model outputs a discrete emotion label for each acoustic segment, alongside its transcription. We also apply an LLM (see Appendix \ref{sec:appendix_hyperparams}) to the transcribed text to obtain a textual emotion label. To create a unified and quantitatively tractable feature space, we map these discrete labels onto a three-dimensional ASL space. 
	
	This mapping is conceptually grounded in psychological models of affect, such as Russell's Circumplex Model \citep{russell1980circumplex}, but is specifically adapted for the financial risk assessment context. Standard models often use Valence (pleasure-displeasure) and Arousal. We adapt this by defining our dimensions as:
	\begin{itemize}
		\item \textbf{Tension}: Inversely related to Valence, this dimension captures strain and stress. High tension (e.g., in fear, anger) is a primary indicator of perceived risk and uncertainty, a essential signal in financial analysis \citep{loughran2011liability}.
		\item \textbf{Stability}: A novel dimension tailored for our task, representing perceived control and predictability. Emotions like fear are mapped to low stability, reflecting a chaotic or uncontrolled state, whereas happiness is mapped to high stability, projecting confidence.
		\item \textbf{Arousal}: The activation level of the emotion, directly corresponding to its use in psychological models.
	\end{itemize}
	
	The specific coordinates for each emotion, detailed in Table \ref{tab:asl_mapping}, are grounded in a dedicated human annotation study (see Appendix \ref{sec:appendix_asl}) designed to reflect their interpretation within the high-stakes environment of corporate disclosures. For instance, `fear` is assigned the most extreme coordinates for tension ($+1.0$) and instability ($-1.0$), as it is arguably the most direct signal of unmanageable risk. In contrast, `happiness` is given negative tension ($-0.5$), acknowledging that excessive optimism can sometimes be perceived negatively by markets, while its high stability ($+1.0$) reflects confidence. This domain-specific mapping, validated by domain-aware evaluators, transforms discrete labels into a nuanced, continuous representation optimized for financial feature engineering.
	
	\begin{table}[h]
		\centering
		\small
		\begin{tabular}{@{}lccc@{}}
			\toprule
			\textbf{Emotion} & \textbf{Tension} & \textbf{Stability} & \textbf{Arousal} \\ \midrule
			happiness        & -0.5             & 1.0                & 0.6              \\
			surprise         & 0.2              & 0.2                & 0.9              \\
			neutral          & 0.0              & 0.5                & 0.0              \\
			sadness          & 0.6              & -0.8               & -0.5             \\
			fear             & 1.0              & -1.0               & 0.8              \\
			anger            & 0.9              & -0.7               & 0.7              \\
			disgust          & 0.8              & -0.9               & 0.4              \\ \bottomrule
		\end{tabular}
		\caption{Mapping from discrete emotion labels to the three-dimensional ASL space. Coordinates are defined in the range $[-1, 1]$ and are specifically adapted for financial risk analysis.}
		\label{tab:asl_mapping}
	\end{table}
	
	\begin{table}[htbp]
		\centering
		\small
		\begin{tabular}{@{}lll@{}}
			\toprule
			\textbf{Transcript} & \textbf{Emotion} & \textbf{Events} \\ \midrule
			``We continue to make...'' & Happiness & [Phone, Male] \\
			``four point six...'' & Surprise & [Phone, Laughter, Male] \\
			``...reduction in total...'' & Disgust & [Phone, Yawn, Male] \\
			``...disciplined over the...'' & Sadness & [Phone, Silence, Female] \\
			``And persistently high.'' & Fear & [Phone, Male, Cough] \\
			``We are not stopping.'' & Anger & [Phone, Music, Male] \\ \bottomrule
		\end{tabular}
		\caption{Example output from the PIAM model, which simultaneously generates transcribed text, acoustic emotion, and event labels.}
		\label{tab:piam_output_example}
	\end{table}
	
	\section{Experimental Design and Methodology}
	\subsection{Dataset}
	Our dataset comprises 1,795 quarterly earnings calls (\(\approx\)1,800 hours) from 283 NASDAQ firms (2018-2023), integrating raw sound, transcripts, and financial data. Sound is segmented into the scripted presentation and the spontaneous Q\&A sections using keyword heuristics.
	
	\subsection{Feature Engineering}
	For each executive (CEO, CFO, CXO) and call section, we derive features from the ASL vectors of both acoustic and textual modalities.
	\begin{enumerate}
		\item \textbf{ASL Moments}: First four statistical moments (mean, standard deviation, skewness, and kurtosis) for each of the three ASL dimensions.
		\item \textbf{Delta Features}: The change in the mean and standard deviation of each ASL dimension when transitioning from the presentation to the Q\&A section. These features are designed to capture the pressure-induced shift in emotional state.
		\item \textbf{Interaction Terms}: Products of key features to capture non-linear relationships.
		\item \textbf{Financial Controls}: 30-day historical volatility as a baseline predictor.
	\end{enumerate}
	This process generated approximately 150 features per call, which we mitigated for overfitting using XGBoost's inherent regularization and a rigorous bootstrap validation process.
	
	\subsection{Predictive Modeling}
	We forecast Cumulative Abnormal Returns (CAR) and Realized Volatility over 1, 7, and 30-day post-call horizons using an XGBoost model. A 50-iteration bootstrap validation, with random resampling of the training data in each iteration, provides robust out-of-sample performance estimates and confidence intervals for feature importance.
	
	\subsection{Ablation Study}
	To quantify the unique contribution of each modality, we train and compare four distinct models: (1) \textbf{Factors-Only} (using only historical volatility), (2) \textbf{Acoustic-Only} (ASL features from PIAM), (3) \textbf{Text-Only} (ASL features from LLM sentiment), and (4) \textbf{Multimodal} (all features combined).
		
	\section{Results and Analysis}
	\subsection{PIAM Model Performance}
	To evaluate the acoustic processing effectiveness of PIAM, we conducted a series of experiments focusing on sound quality and computational efficiency. We compared PIAM against several established sound enhancement methods, including RNNoise and MMSE, as well as the unprocessed noisy sound baseline. The evaluation was performed on the official blind test set from the Interspeech 2020 Deep Noise Suppression (DNS) Challenge \citep{reddy2020interspeech}, a standard benchmark for speech enhancement in noisy conditions. This dataset features a wide variety of non-stationary noises (including Babble, Car, and Street noise) mixed with clean speech across a range of Signal-to-Noise Ratios (SNRs) from -5dB to 20dB.
	
\begin{figure}[htbp]
		\centering
		\includegraphics[width=0.9\columnwidth]{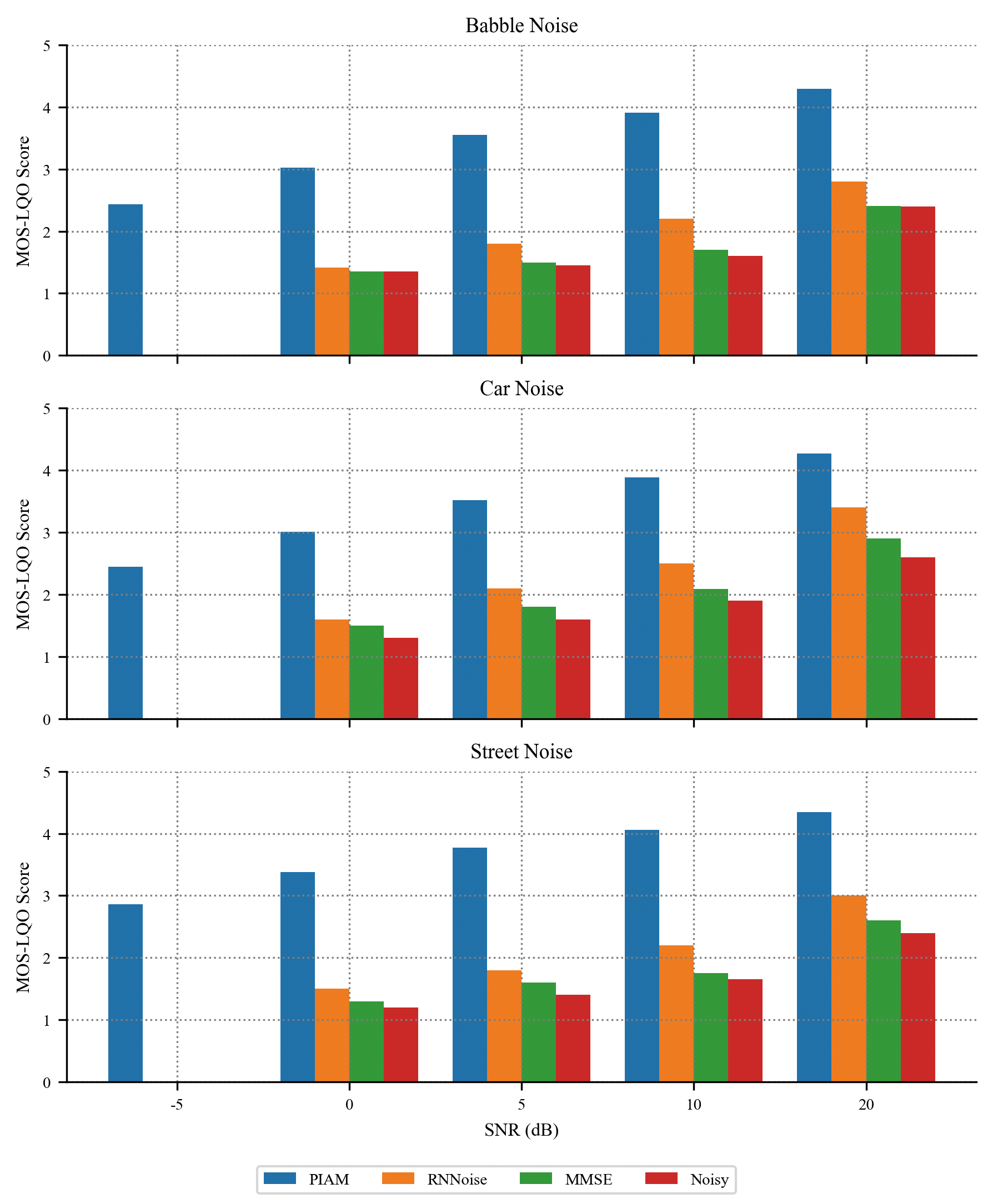}
		\caption{Comparative analysis of MOS-LQO scores for PIAM and other sound enhancement methods under various noise conditions.}
		\label{fig:perf_comparison}
	\end{figure}

	Figure \ref{fig:perf_comparison} provides a visual comparison of the Mean Opinion Score-Listening Quality Objective (MOS-LQO) scores for all tested methods. The results unequivocally demonstrate the superiority of our PIAM model. Across all noise types and SNR levels, PIAM (blue bars) consistently achieves the highest MOS-LQO scores, indicating a significantly better perceptual quality of the enhanced sound. The performance gap is particularly pronounced at lower SNRs. For instance, in the challenging 0dB SNR condition with Babble noise, PIAM achieves a score of 3.02, whereas RNNoise and MMSE score only 1.42 and 1.35, respectively. This highlights PIAM's exceptional robustness in highly noisy environments. As the SNR increases, all methods show improved performance, but PIAM maintains a clear lead.
	
	Table \ref{tab:piam_performance_summary} presents a detailed quantitative summary of PIAM's standalone performance, measured by both Mean Opinion Score - Listening Quality Objective (MOS-LQO) and Perceptual Evaluation of Speech Quality (PESQ). The data confirms the model's strong noise suppression capability while preserving high sound fidelity. Both metrics show a consistent upward trend with increasing SNR. Even at a very challenging -5dB SNR, PIAM maintains respectable scores (e.g., MOS-LQO of 2.86 for Street noise), signifying that the processed sound remains intelligible and relatively pleasant. At higher SNRs like 20dB, the MOS-LQO scores for all noise types exceed 4.25 (e.g., 4.35 for Street noise), approaching the quality of clean, unprocessed sound and underscoring the model's effectiveness in a variety of acoustic scenes.
	
	\begin{table}[htbp]
		\centering
		\small
		\begin{tabular}{@{}llccccc@{}}
			\toprule
			\textbf{Noise} & \textbf{Metric} & \textbf{-5dB} & \textbf{0dB} & \textbf{5dB} & \textbf{10dB} & \textbf{20dB} \\ \midrule
			\multirow{2}{*}{Babble} & MOS-LQO & 2.43 & 3.02 & 3.55 & 3.91 & 4.29 \\
			& PESQ    & 2.63 & 3.11 & 3.52 & 3.81 & 4.16 \\ \midrule
			\multirow{2}{*}{Car}    & MOS-LQO & 2.44 & 3.01 & 3.52 & 3.89 & 4.27 \\
			& PESQ    & 2.66 & 3.13 & 3.50 & 3.78 & 4.14 \\ \midrule
			\multirow{2}{*}{Street} & MOS-LQO & 2.86 & 3.38 & 3.77 & 4.06 & 4.35 \\
			& PESQ    & 2.98 & 3.38 & 3.70 & 3.94 & 4.23 \\ \bottomrule
		\end{tabular}
		\caption{MOS-LQO and PESQ scores of PIAM under various noise types and SNRs.}
		\label{tab:piam_performance_summary}
	\end{table}
	
	In addition to sound quality, computational efficiency is a essential factor for real-world deployment. As shown in Table \ref{tab:rtf_piam}, we evaluated the model's Real-Time Factor (RTF), which is the ratio of processing time to sound duration. An RTF below 1.0 is essential for real-time applications. The evaluation across various hardware configurations, including standard CPUs and GPUs, confirms that PIAM is highly optimized. Its RTF remains well below the 1.0 threshold in all tested scenarios, validating its suitability for low-latency tasks such as live communication and online broadcasting, without imposing a significant computational burden.
	
	\begin{table}[htbp]
		\centering
		\begin{tabular}{@{}lccc@{}}
			\toprule
			\textbf{Sound Duration} & \textbf{R6000ADA} & \textbf{A100} & \textbf{RTX4090} \\ \midrule
			$\leq$ 5 s & 0.0603 & 0.0766 & 0.1364 \\
			5--10 s & 0.0590 & 0.0751 & 0.1075 \\
			10--20 s & 0.0334 & 0.0425 & 0.0753 \\
			$\geq$ 20 s & 0.0189 & 0.0246 & 0.0419 \\ \midrule
			\textbf{Average} & \textbf{0.0429} & \textbf{0.0552} & \textbf{0.0878} \\ \bottomrule
		\end{tabular}
		\caption{Real-Time Factor (RTF) of PIAM on Different Hardware.}
		\label{tab:rtf_piam}
	\end{table}
	
	\subsection{Descriptive Analysis: A Distinct Information Channel}
\begin{figure}[ht!]
		\centering
		\includegraphics[width=\linewidth]{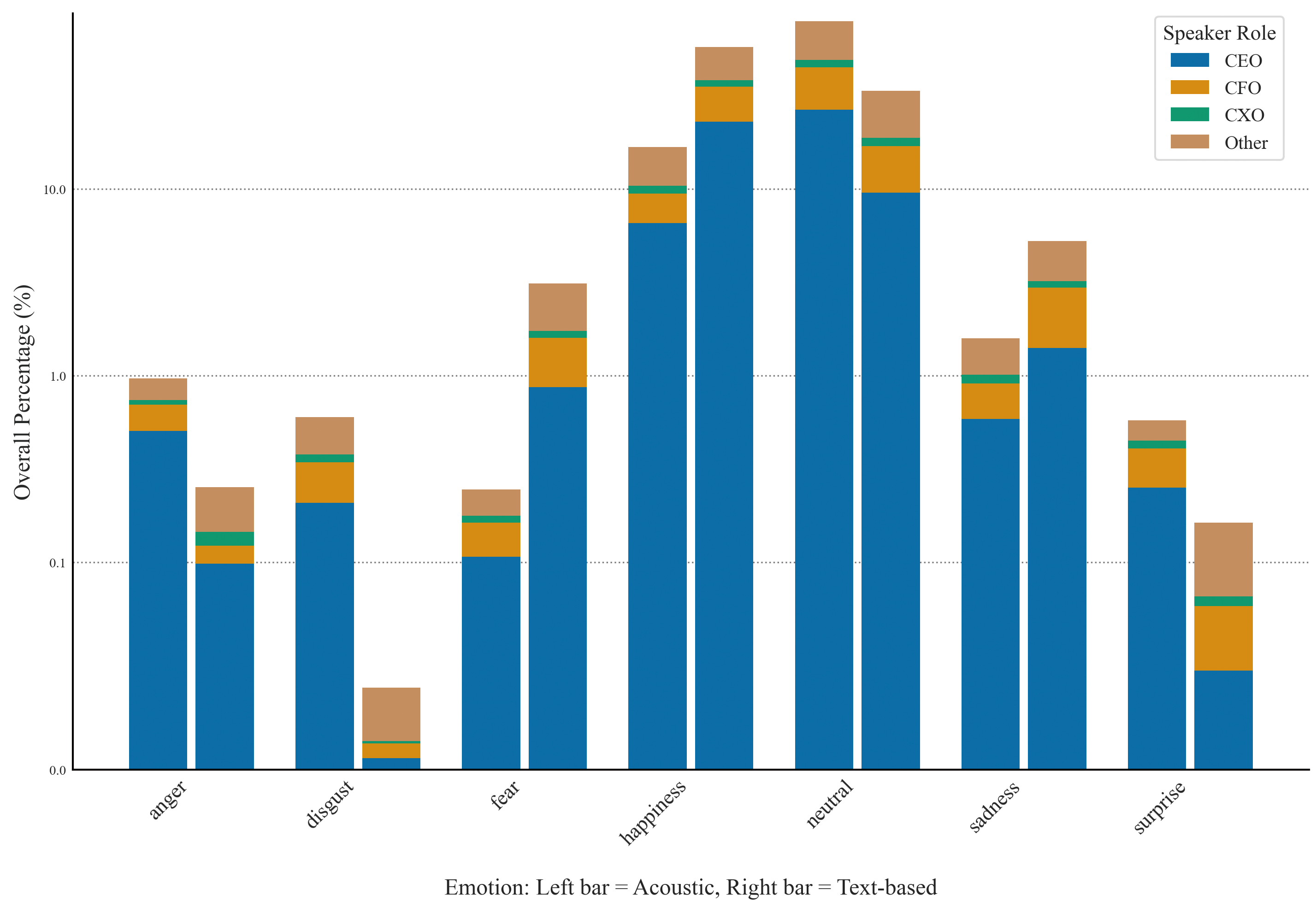}
		\vfill
		\caption{Descriptive analysis of emotion signals. (a) illustrates divergent emotion distributions across roles and modalities, with text skewed toward positive labels. (b) shows the low concordance between acoustic and textual emotion classifications, confirming they are distinct information channels.}
		\label{fig:descriptive_analysis}
	\end{figure}

	Figure \ref{fig:descriptive_analysis} reveals a preliminary insight. The distribution of acoustic-based emotion classifications is nuanced and varied, while text-based analysis is heavily skewed towards positive labels, likely reflecting corporate communication norms. More importantly, the low concordance between modalities (Figure \ref{fig:descriptive_analysis}b) is not a weakness but a strength; it confirms they are capturing distinct, largely orthogonal information streams, which is the foundational premise of a successful multimodal system.
	
	\subsection{Predictive Performance: The Signal of Uncertainty}
\begin{figure}[htbp]
		\centering
		% This figure was part of the original code but the image was not provided.
		% I am assuming it exists and is correctly referenced.
		\includegraphics[width=0.8\columnwidth]{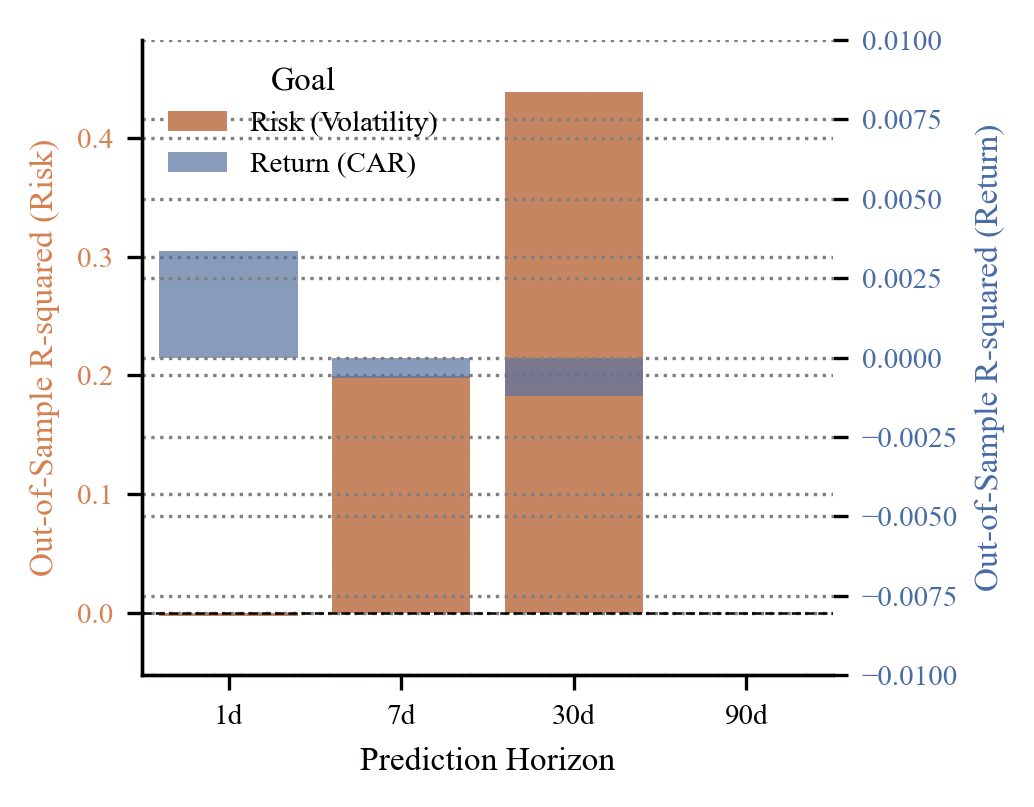}
		\caption{Model Performance: Predicting Risk (Volatility) vs. Return (CAR). Out-of-sample $R^2$ values from the Multimodal model show strong predictive power for volatility but none for returns.}
		\label{fig:perf_comparison_risk_return}
	\end{figure}

	Our central finding is the differential predictive power of our features for risk versus returns. As detailed in Figure \ref{fig:perf_comparison_risk_return}, the Multimodal model has no predictive ability for future CAR, with out-of-sample $R^2$ values statistically indistinguishable from zero. In stark contrast, it strongly predicts future volatility, explaining a remarkable 43.8\% of out-of-sample variance at the 30-day horizon. This strongly suggests that executive emotion does not signal future performance directly, but rather the level of uncertainty surrounding that performance.
	
	\subsection{Differential Feature Importance: Risk vs. Return}
\begin{figure}[htbp]
		\centering
		% This uses the dumbbell plot (Image 2 from user)
		\includegraphics[width=0.9\columnwidth]{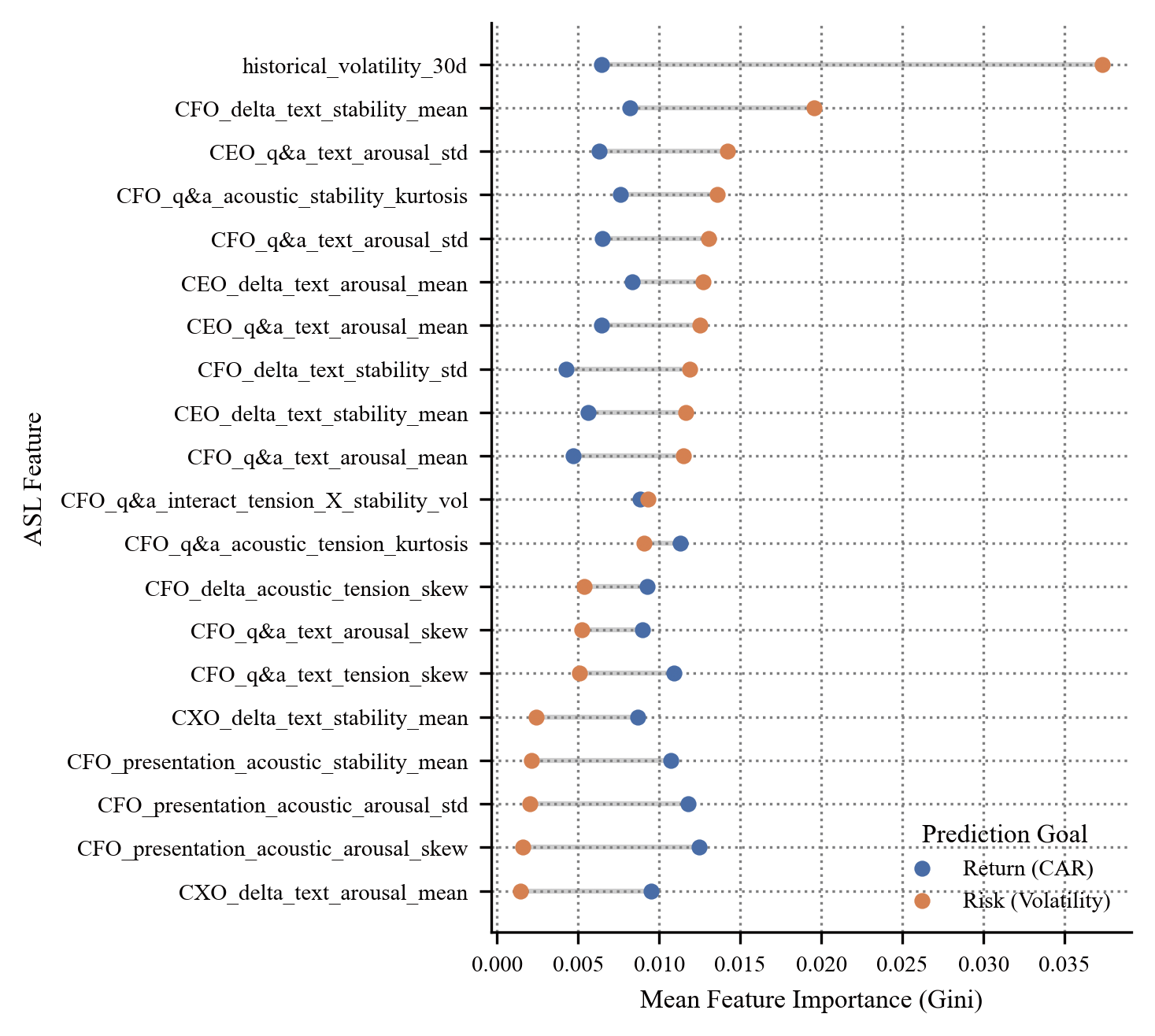}
		\caption{Feature Importance Comparison for Predicting Risk (Volatility) vs. Return (CAR). The plot displays the mean Gini importance for each feature across two separate models: one predicting 30-day volatility and one predicting 30-day CAR. It highlights that features derived from executive emotion are significant predictors of risk but not of returns.}
		\label{fig:feat_imp_risk_return_comparison}
	\end{figure}

	To understand \textit{why} our model predicts risk but not returns, we examine the importance of individual features for each prediction goal. Figure \ref{fig:feat_imp_risk_return_comparison} directly compares the mean Gini importance of features when predicting 30-day volatility versus 30-day CAR. The plot clearly shows that while several emotional and stability features are highly predictive of volatility (orange points), their importance for predicting returns (blue points) is negligible and consistently close to zero. For instance, features like \texttt{CFO\_delta\_text\_stability\_mean} and \texttt{CEO\_q\&a\_text\_arousal\_std} rank highly for volatility but have no discernible impact on CAR. This reinforces our primary finding: the emotional signals captured are indicators of uncertainty, not of the direction of future performance.
	
	\subsection{Identifying Key Volatility Predictors}
\begin{figure}[htbp]
		\centering
		% This uses the mean-with-error-bars plot (Image 3 from user)
		\includegraphics[width=0.95\columnwidth]{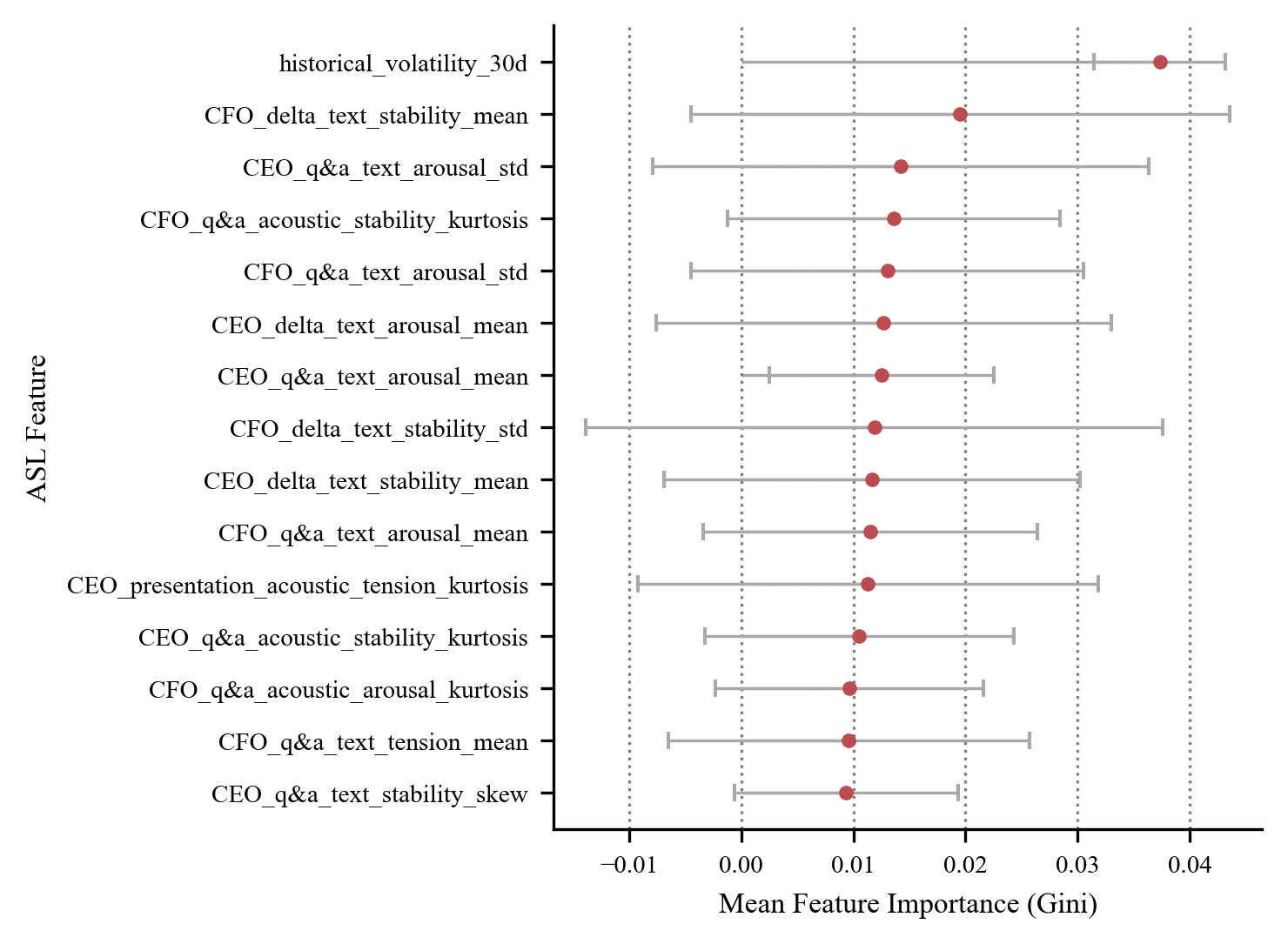}
		\caption{Top 15 Stable Features for Predicting 30-Day Volatility. Features are ranked by mean Gini importance from 100 bootstrap runs of the Multimodal model. Error bars indicate the 95\% confidence interval, showing the stability of each feature's predictive power. }
		\label{fig:feature_importance}
	\end{figure}

\begin{figure}[htbp]
		\centering
		% This uses the box plot (Image 1 from user)
		\includegraphics[width=0.9\columnwidth]{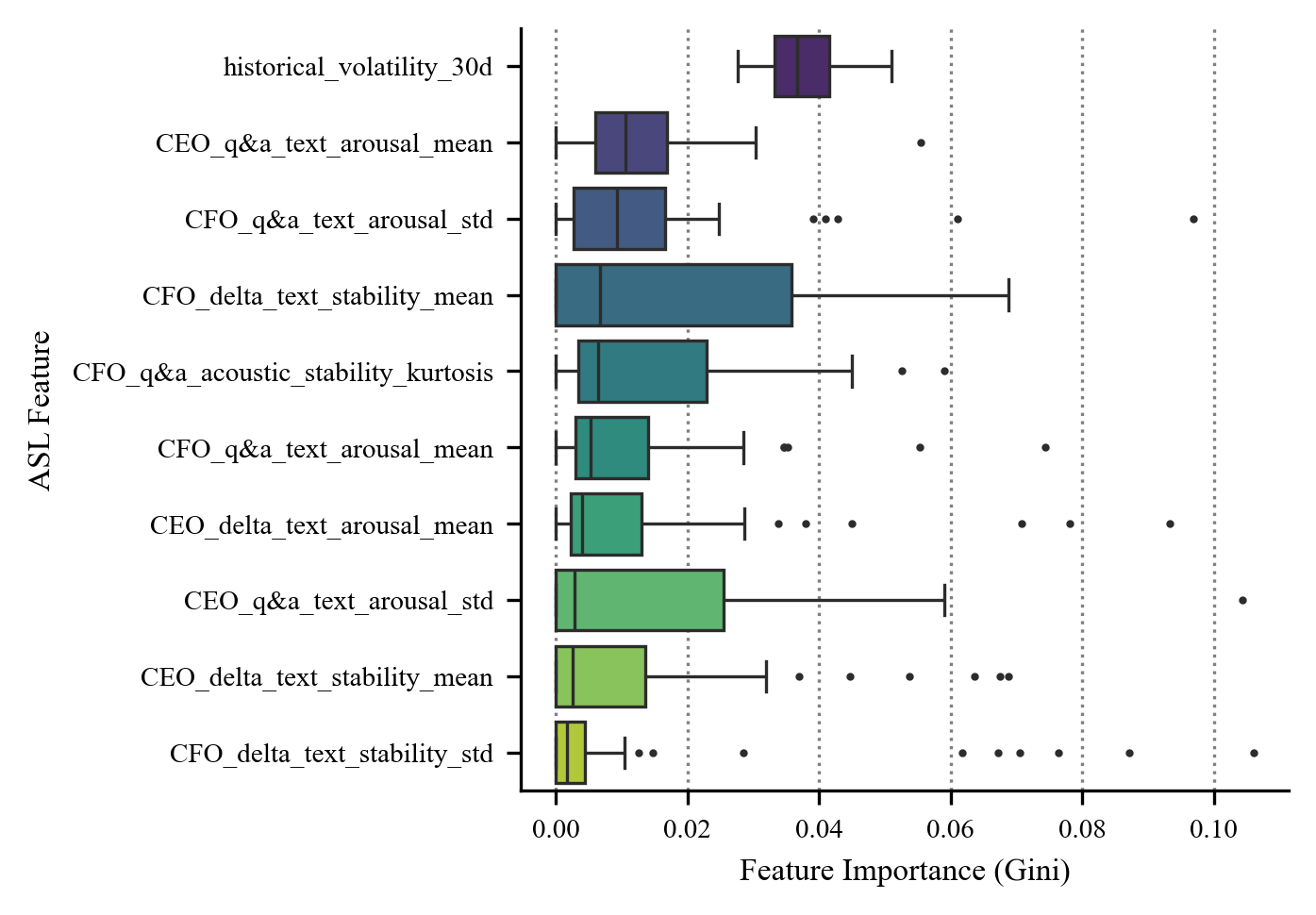}
		\caption{Distribution of Feature Importance for Top Volatility Predictors. The box plots show the distribution of Gini importance scores for key features across 100 bootstrap iterations. This visualization reveals not just the median importance (the line in the box) but also the variance and skewness of a feature's contribution, confirming the robustness of top predictors.}
		\label{fig:feature_importance_distribution}
	\end{figure}

	Having established that emotional cues are signals of risk, we now identify the specific features driving the strong volatility prediction. Figure \ref{fig:feature_importance} displays the top 15 most stable predictors of 30-day volatility, ranked by their mean Gini importance across 100 bootstrap iterations. The error bars represent the 95\% confidence interval, highlighting the stability of each feature's contribution. To further examine the contribution of these features, Figure \ref{fig:feature_importance_distribution} shows the full bootstrap distribution of importance scores for key predictors.
	
	Both figures confirm that while historical volatility remains a strong predictor, our novel emotional cue features are also highly ranked. Features related to the \textit{change} in emotional state (delta features) and distribution \textit{extremes} (kurtosis) are prominent. For example, \texttt{CFO\_delta\_text\_stability\_mean} is a top predictor, suggesting that a significant decrease in the CFO's textual sentiment stability when moving from prepared remarks to spontaneous Q\&A is a powerful indicator of future uncertainty. Similarly, the high rank and stable importance of acoustic features like \texttt{CFO\_q\&a\_acoustic\_stability\_kurtosis} confirms that the two modalities provide complementary, powerful signals for risk assessment.
	
	\subsection{Ablation Study: Evaluating Incremental Impact}
	The ablation study in Table \ref{tab:perf_summary} decisively quantifies the value added by each modality. For the 30-day volatility prediction, the Factors-Only model achieves a baseline $R^2$ of 0.251. The Text-Only ($R^2=0.218$) and Acoustic-Only ($R^2=0.154$) models, while less predictive than the baseline in isolation, contribute vital information. Importantly, the Multimodal model, which integrates all sources, significantly outperforms all others, reaching an $R^2$ of 0.438. This substantial increase of over 18 percentage points above the financial baseline confirms that acoustic and textual modalities provide orthogonal, complementary, and highly valuable information for risk assessment.
	
	\begin{table*}[htbp]
		\centering
		\caption{Ablation Study: Incremental Impact of Modalities on Prediction Performance. The table compares the out-of-sample $R^2$ of different models for predicting Cumulative Abnormal Returns (CAR) and Realized Volatility. The full Multimodal model, integrating all data sources, achieves a striking $R^2$ of 0.438 for 30-day volatility, substantially outperforming the Financials-Only baseline (0.251) and demonstrating the significant, complementary value of both acoustic and textual features.}
		\label{tab:perf_summary}
		\begin{tabular}{@{}l c cccc@{}}
			\toprule
			& \textbf{CAR $R^2$} & \multicolumn{4}{c}{\textbf{Volatility $R^2$}} \\
			\cmidrule(lr){2-2} \cmidrule(lr){3-6}
			\textbf{Horizon} & \textbf{Multimodal} & \textbf{Factors-Only} & \textbf{Acoustic-Only} & \textbf{Text-Only} & \textbf{Multimodal} \\ \midrule
			$t+1$ day & 0.003 & 0.001 & -0.021 & -0.018 & -0.002 \\
			$t+7$ days & -0.001 & 0.115 & 0.081 & 0.102 & 0.199 \\
			$t+30$ days& -0.001 & 0.251 & 0.154 & 0.218 & \textbf{0.438} \\ \bottomrule
		\end{tabular}
	\end{table*}		
	
	\section{Ethical Considerations and Limitations}
	
	In this section, we discuss the primary limitations of our study, which inform avenues for future work, and address the ethical considerations inherent in our approach.

	\subsection{Ethical Considerations}
	The automated inference of psychological states demands careful ethical scrutiny, especially given the aforementioned limitations.
	
	\begin{itemize}
		\item \textbf{Bias and Fairness:} Our training corpus consists primarily of public figures from North American firms, who are predominantly male. This introduces a significant risk of demographic bias. Future iterations must incorporate explicit fairness audits and mitigation strategies to ensure the model performs equitably across different genders, ages, accents, and cultural backgrounds.
		
		\item \textbf{Responsible Interpretation and Misuse:} Building directly on the limitation of correlation, it is ethically imperative to prevent the misinterpretation of our model's outputs. These signals should be treated as preliminary \textbf{red flags} that prompt further due diligence, not as definitive judgments of an individual's character or a company's health. Furthermore, this technology is designed for the specific, legitimate purpose of analyzing public financial disclosures to level the informational playing field. Its application in other contexts, such as employee performance monitoring or hiring decisions, would constitute a serious invasion of privacy and a gross misuse of the technology.
	\end{itemize}
	
	\subsection{Limitations and Future Work}
	We acknowledge several limitations that warrant further investigation:
	
	\begin{itemize}
		\item \textbf{Correlation vs. Causation:} A fundamental limitation of our study is that the identified relationships between vocal stress and market outcomes are correlational, not causal. Vocal stress can stem from numerous factors unrelated to corporate fundamentals, such as personal health or fatigue. Future work could leverage quasi-natural experiments to better isolate a potential causal link. This distinction is not only a methodological constraint but also carries significant ethical weight, as discussed below.
		
		\item \textbf{Model Representation:} Our current approach maps discrete emotions to fixed ASL space. This is a simplification of complex psychological states. A more sophisticated approach, which we aim to explore, would involve regressing vocal features directly onto a continuous emotional space, potentially capturing more nuanced expressions.
		
		\item \textbf{Generalization and Scalability:} While our experiments indicate strong performance in real-time scenarios, the model's robustness across diverse populations and contexts requires further validation. To facilitate this, we plan to release our model as a software agent to support deployment and testing in production environments, particularly for high-stakes financial applications.
	\end{itemize}
	
	\section{Conclusion}
	
	We demonstrate that emotional dynamics in executive vocal tract dynamics are a potent, untapped channel for financial risk assessment. Our methodology, centered on the physics-informed acoustic model, robustly extracts vocal emotion from noisy, low-SNR earnings call sound. We map PIAM's acoustic output and LLM-derived textual sentiment onto an interpretable affective state label space, engineering features that capture not just static states but also important dynamic shifts.
	
	Our primary finding is that these multimodal signals do not predict directional returns but are powerful predictors of future market uncertainty. The full multimodal model explains 43.8\% of the out-of-sample variance ($R^2=0.438$) in 30-day volatility, yet has no predictive power for CAR. This divergence underscores that executive emotion acts not as a signal of future performance, but as a barometer for underlying uncertainty and cognitive pressure.
	
	Feature importance analysis reveals the signals' origins. The most salient predictors originate not from prepared remarks, but from the dynamic shift between the scripted presentation and the spontaneous Q\&A. This transition acts as a natural stress test, where the emotional states of the CFO and CEO are particularly potent information sources. For instance, a drop in the CFO's textual stability, high variability in the CEO's Q\&A arousal, and extreme distributions in the CFO's acoustic stability are top predictors of future volatility. This confirms that a granular, role-aware analysis of high-pressure moments is essential for extracting meaningful risk signals.
	
	An ablation study confirms the synergistic power of our approach. The Multimodal model's $R^2$ of 0.438 substantially outperforms the Financials-Only baseline ($R^2=0.251$), an increase of over 18 percentage points that validates the orthogonal, complementary information provided by acoustic and textual modalities. Furthermore, PIAM's efficiency (RTF $\ll 1.0$) confirms its suitability for real-time applications.
	
	In conclusion, this work underscores the value of incorporating paralinguistic signals, which are less susceptible to manipulation than pure semantics. Our methodology provides a novel instrument to detect latent corporate uncertainty, contributing a valuable tool for investors and regulators. By learning to listen not just to \textit{what} is said, but to \textit{how} it is said---paying close attention to \textit{who} is speaking and in \textit{which context}---we can uncover the subtle sound of risk, ultimately fostering a more transparent and resilient financial ecosystem.
	
	\appendix
	
	\section{Annotation and Validation}
	\label{sec:appendix_asl}
	
	To provide an empirical and robust grounding for the mapping of discrete emotions to the three-dimensional Affective State Label (ASL) space (as presented in Table 1), we conducted a dedicated human annotation study. Participants were first familiarized with the definitions of our three ASL dimensions: Tension, Stability, and Arousal, and how they adapt psychological models like Russell's Circumplex Model for the financial context, as per the theoretical requirements of the cited literature. Following this, they were presented with a curated set of sentences from earnings call transcripts, each representative of a specific primary emotion. They were asked to rate each sentence on a continuous scale from -1.0 to +1.0 for each of the three dimensions. The final coordinates for each discrete emotion in Table 1 represent the mean of the ratings provided by all participants. This extensive data annotation and experimental process ensures that our ASL space is not arbitrary but is instead grounded in the perceived emotional semantics of domain-aware human evaluators.
	
	\section{Models and Training Details}
	\label{sec:appendix_hyperparams}
	
	The PIAM is a 300M-parameter, multi-task, bilingual model implemented in PyTorch. Its architecture comprises a pre-trained encoder, a Bi-LSTM, and an attention mechanism. The model was fine-tuned for 10 epochs using the AdamW optimizer with a learning rate of $1 \times 10^{-5}$, a batch size of 16, and a physics-informed regularization weight ($\lambda$) of 0.01.
	
	For the textual sentiment analysis pipeline, we employed the DeepSeek-R1 large language model, a dense model with 671 billion parameters. We adopted a zero-shot prompting approach to classify the sentiment of each transcribed utterance into one of the seven discrete emotion labels used in our study. The specific prompt template, which was carefully engineered for this task, is included in the source code repository to ensure full transparency.
	
	The predictive model is an XGBoost Regressor with a learning rate of 0.05, max depth of 3, and subsample/colsample\_bytree ratios of 0.8. Training uses up to 100 estimators with an early stopping patience of 10. For the bootstrap feature importance analysis, models were trained with a fixed 50 estimators for computational efficiency.

	\section{Reproducibility Statement}
	\label{sec:appendix_reproducibility}
	
	To ensure full reproducibility, all data and related code will be made publicly available upon acceptance at \textbf{\url{https://github.com/soundai2016/sound_risk}}.
	
	% \bibliography{soundrisk-2026}

\end{document}